\documentclass[letterpaper, 10 pt, conference]{ieeeconf}  % Comment this line out if you need a4paper
\IEEEoverridecommandlockouts
\usepackage{graphics} % for pdf, bitmapped graphics files
\usepackage{graphicx} 
\usepackage{amssymb}

\title{\LARGE \bf
Multi-Channel Neural Network for Assessing Neonatal Pain from Videos}

\author{Md Sirajus Salekin$^{1}$, Ghada Zamzmi$^{1,3}$, Dmitry Goldgof$^{1}$, Rangachar Kasturi$^{1}$, Thao Ho$^{2}$, and Yu Sun$^{1}$% <-this % stops a space
\thanks{$^{1}$Department of Computer Science and Engineering, University of South Florida, FL, USA}%
\thanks{$^{2}$College of Medicine Pediatrics, USF Health, University of South Florida, FL, USA}%
\thanks{$^{3}$The current affiliation of Dr. Zamzmi is the National Library of Medicine, National Institutes of Health, MD, USA}%
}

\begin{document}

%%%%%%%%%%%%%%%%%%%%%%%%%%%%%%%%%%%%%%%%%%%%%%%%%%%%%%%%%%%%%%%%%%%%%%%%%%%%%%%%
\onecolumn
\noindent \huge IEEE Copyright Notice\\

\noindent \normalsize \textcopyright \hspace{0.05cm} 2019 IEEE. Personal use of this material is permitted. Permission from IEEE must be obtained for all other uses, in any current or future media, including reprinting/republishing this material for advertising or promotional purposes, creating new collective works, for resale or redistribution to servers or lists, or reuse of any copyrighted component of this work in other works.\\

\noindent \large {Accepted to be Published in: Proceedings of the 2019 IEEE International Conference on Systems, Man, and Cybernetics (IEEE SMC 2019), October 06-09, 2019, Bari, Italy.}
\twocolumn
\normalsize 
%%%%%%%%%%%%%%%%%%%%%%%%%%%%%%%%%%%%%%%%%%%%%%%%%%%%%%%%%%%%%%%%%%%%%%%%%%%%%%%%

\maketitle
\thispagestyle{empty}
\pagestyle{empty}

%%%%%%%%%%%%%%%%%%%%%%%%%%%%%%%%%%%%%%%%%%%%%%%%%%%%%%%%%%%%%%%%%%%%%%%%%%%%%%%%

\begin{abstract}
Neonates do not have the ability to either articulate pain or communicate it non-verbally by pointing. The current clinical standard for assessing neonatal pain is intermittent and highly subjective. This discontinuity and subjectivity can lead to inconsistent assessment, and therefore, inadequate treatment. In this paper, we propose a multi-channel deep learning framework for assessing neonatal pain from videos. The proposed framework integrates information from two pain indicators or channels, namely facial expression and body movement, using convolutional neural network (CNN). It also integrates temporal information using a recurrent neural network (LSTM). The experimental results prove the efficiency and superiority of the proposed temporal and multi-channel framework as compared to existing similar methods. 
\end{abstract}

%%%%%%%%%%%%%%%%%%%%%%%%%%%%%%%%%%%%%%%%%%%%%%%%%%%%%%%%%%%%%%%%%%%%%%%%%%%%%%%%

\section{Introduction}
Pain is an unpleasant emotional experience that can be expressed verbally (adults) or non-verbally (children). Unfortunately, neonates do not have the ability to either articulate pain or communicate it non-verbally by pointing. Pain in neonates occurs as different behavioral and physiological cues. By monitoring these cues, caregivers can understand neonates' pain and develop suitable treatments. The caregivers' observation of neonatal pain is a tedious job that requires significant time and resources. Moreover, it is intermittent and depends entirely on the subjective judgment of the observer. 

The discontinues assessment of pain may lead to missing the pain during the postoperative period when the neonate is left unattended (under-treatment). The observers' subjectivity can result in inconsistent assessment, which can lead to inconsistent treatment. The inadequate treatment (under- or over-treatment) of neonatal pain can cause serious damages to the brain of neonates as discussed thoroughly in \cite{puchalski2002reality, ranger2007current}. Hence, developing automated methods that continuously monitor pain is needed in all Neonatal Intensive Care Unit (NICU) to provide a standardized and continuous assessment. 

Currently, neonatal pain is measured using clinical pain scales (e.g., NIPS \cite{hudson2002validation} and NPASS \cite{hummel2008clinical}) that provide a score for different pain cues followed by adding all the scores together to obtain the final level of pain. Examples of these cues include facial expression, body movement, crying sound, and vital sign readings. Among all pain cues, facial expression is considered the most validated, prominent, and pain-specific cue \cite{grunau1987pain,peters2003neonatal,zhi2018infants}. 

\begin{figure}[t]
		\begin{center}
		\includegraphics[width=0.28\textwidth]{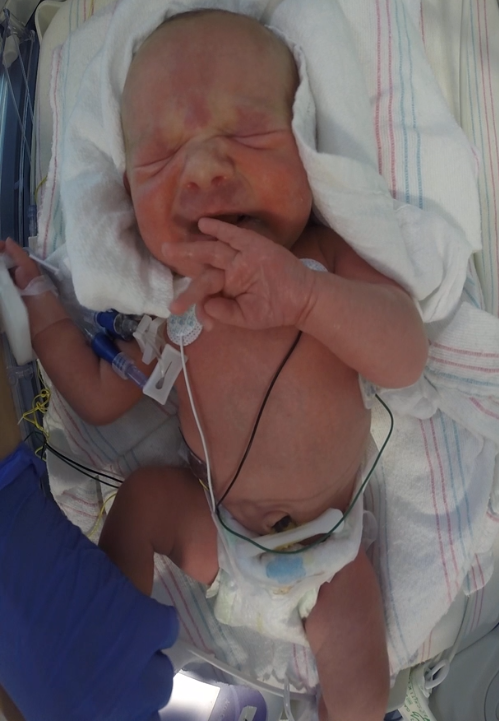}
		\caption{Facial and bodily pain stimulus of neonate}
		\label{fig_painFaceBody}
		\end{center}
	\end{figure}

Although facial expression is considered the most pain-specific indicator, extracting features from facial expression and other indicators is necessary for two main reasons. First, combining multiple pain indicators allows to assess pain in case of missing data (occluded face). Second, there is a strong correlation between different pain indicators \cite{holsti2005body,bellieni2012pain}. An illustration of this correlation is depicted in Figure \ref{fig_painFaceBody}. As it can be seen from the Figure \ref{fig_painFaceBody}, the pain stimulus triggers both facial expression and body movement. Due to these reasons, we think it is important to assess neonatal pain using a shared representation of different pain indicators (e.g., facial expression and body movement). 

Existing automated methods for assessing pain of neonates from facial expression are broadly classified into handcrafted methods and deep learning methods \cite{zamzmi2018review}. 

Most existing methods fall under the first category. The first handcrafted method for neonatal pain assessment from facial expression was introduced by Brahnam et al. \cite{brahnam2007introduction}. In this work \cite{brahnam2007introduction}, a neonatal facial dataset named COPE was introduced and used to assess pain from facial features extracted from the images. The extracted images' features are used with PCA, LDA, and SVM classifiers to classify the static images as pain or no-pain. COPE dataset has 204 static images of 26 healthy infants. Nanni et al. \cite{nanni2010local} used different variations of Local Binary Patterns (LBP) descriptor to classify images of COPE dataset as pain or no-pain. Instead of detecting the label of pain using a binary classifier, Gholami et al. \cite{gholami2010relevance} used Relevance Vector Machine (RVM) to detect pain and estimate the intensity of the detected pain. The intensity estimation was determined using the posterior probability provided by the RVM. 

Zamzmi et al. \cite{zamzmi2016approach} proposed a multimodal neonatal pain analysis system. They used a decision fusion after getting the predictions from multiple modalities; i.e. facial expression, body movement, crying sound and vital sign. The prediction for each of these modalities was obtained using binary classifiers (e.g., SVM) trained using handcrafted features. Recently, deep learning methods \cite{rodriguez2017deep, bellantonio2016spatio} have become popular in pain classification. In case of infants, Celona and Manoni \cite{celona2017neonatal} proposed a framework that combines both handcrafted and deep features for classifying COPE images as pain/no-pain. Specifically, they combined LBP, HOG, pre-trained VGG-face and Mapped LBP+CNN features to represent the final feature vector. The extracted feature vector was used to train SVM to perform binary classification (pain/no-pain). 

There are two main drawbacks of existing automatic methods for assessing neonatal pain. First, the vast majority of the methods that assess pain from facial expression depend highly on facial landmark points for either cropping the face region or extracting features from the face. Facial landmark detection can be challenging in case of neonates since their facial muscles are not well-developed as compared to adults. In addition, neonates hospitalized in the NICU undergo different postoperative care that might lead to complete or partial occlusion of their face (e.g., oxygen mask or tapes). Besides, designing and training landmark points detector is a complex task that requires a relatively large dataset with accurate landmark points annotations. 

Second, these methods, except \cite{zamzmi2016approach}, assess pain from static images. Pain expression is a dynamic event that unfolds in a particular pattern over time. Therefore, it is important to analyze the pain expression over time using temporal methods. In case of adults, various works \cite{rodriguez2017deep, bellantonio2016spatio} reported that LSTM is an effective way for conveying the temporal information of pain expression. The experimental results of Rodriguez et al. \cite{rodriguez2017deep} method, which combined VGG16 CNN with LSTM, showed that using LSTM improves the performance of detecting pain from videos.

In this paper, we propose a novel multi-channel deep neural network framework for detecting neonatal pain from videos. The proposed framework continuously observes two different pain indicators, facial expression and body movement, and feeds them into a convolutional neural network and LSTM. The major contributions of this paper can be summarized as follows: 

\begin{itemize}
    \item We present a multi-channel CNN-based network for detecting pain in neonates. To the best of our knowledge, this paper is the first to explore neonatal pain assessment using a multi-channel CNN network. 
    \item We present a landmark-free approach to assess the neonatal pain from the face and compare this approach to a previous landmark-based approach \cite{zamzmi2016approach}. 
    \item We investigate the correlation between face and body and show how these two pain indicators can be used jointly to provide better representation of pain.
    \item We incorporate recurrent neural network (LSTM) to the multi-channel network to model the temporal changes of neonatal pain expression.
    \item We evaluate the proposed framework on a challenging dataset collected from neonates while they were hospitalized in the NICU. 
    \item We compare the proposed framework's performance with the performance of several handcrafted methods as well as existing CNNs architecture such as VGG-16 and ResNet50. 
\end{itemize}

The rest of this paper is organized as follows. Section II provides background of the CNN and RNN architectures that are utilized to develop our framework. Section III presents our novel framework for assessing neonatal pain using a temporal and multi-channel network. We provide a description of the dataset and the experimental results of the framework in Section IV. Finally, we conclude and discuss future research directions in Section V. 

%%%%%%%%%%%
%%%%%%%%%%%%%%%%%%%%%%%%%%%%%%%%%%%%%%%%%%%%%%%%%%%%%%%%%%%%%%%%%%%%%

\section{Background}
This section presents the architectures of the networks that are used to build the proposed framework. These networks are YOLO, VGG16, and LSTM. 

\subsection{YOLO}
YOLO (You Only Look Once) \cite{redmon2018yolov3} is one of the state-of-the-art real-time object detection algorithms. It is a deep CNN architecture that can effectively: 1) find out the location (bounding box) of a desired object in the entire image and 2) provide the corresponding class label of the detected object with a confidence label. YOLO takes an image as input and divides it into a grid of cells. Each cell predicts a number of bounding boxes in the image. For each bounding box, the model provides relevant confidence about the bounding box as well as the confidence of the object class surrounded by the bounding box. Using a non-maximum suppression techniques, YOLO removes all bounding boxes with low confidence scores and outputs the bounding boxes with the high confidence score. 

Recently, YOLO has become the de facto network for object detection due to its high accuracy and fastness. In this paper, we used two trained YOLO detectors, face detector and body detector, to detect the face and body regions of the neonates. One YOLO was used for neonate's face detection and another one used for neonate's body detection. For face detection, YOLO was trained using the WIDER FACE \cite{yang2016wider} dataset that contains 61 event classes and 32,203 face images. This benchmark face dataset is popular due to its robustness in terms of scale, pose, and occlusion. As for neonate's body detection, we used another YOLO detector, trained on the popular COCO \cite{lin2014microsoft} dataset, which has 80 object categories and 330K images.

\subsection{VGG16}
VGG16 \cite{simonyan2014very} is a well-known CNN architecture for visual classification. VGG16 was the runner-up of the ILSVRC 2014. This network was built by modifying AlexNet architecture \cite{krizhevsky2012imagenet}. Specifically, VGG16 replaces the large kernel-size filters of AlexNet with several $3 \times 3$ kernel-size filters. The main argument was that multiple smaller size kernels can provide multiple non-linear layers that lead to learning more complex features. In VGG16, a series of $3 \times 3$ kernel-size filters are used multiple times to extract more representative features. The network starts with a 64 depth and increases by a factor of 2 after each pooling layer. Finally, 3 fully-connected layers are used before the final classification layer. The first and second fully connected layers have 4096 units and the last layer performs classification using a Softmax function. 

In this paper, two pre-trained VGG16 models are used to extract deep features from neonates' faces. To extract deep features from the neonatal face, we used a VGG16 model trained on the VGGFace2 \cite{cao2018vggface2} dataset. VGGFace2 is the upgrade of previous VGGFace \cite{parkhi2015deep} dataset which is more robust across pose and age. Another pre-trained VGG16 model, trained on popular ImageNet \cite{deng2009imagenet} dataset was used to extract deep features from the entire body of the neonates.

\subsection{LSTM}
Traditional Neural Networks can not capture the temporal changes over time. Hence, RNN (Recurrent Neural Network) was introduced to incorporate temporal information to the traditional Neural Networks. RNN performs prediction or classification using current and past information. The issue of long term dependencies is one limitation of RNN networks. To solve this issue, LSTM (Long Short Term Memory) \cite{hochreiter1997long} was introduced as a special kind of RNN that is able to handle long-term dependency. LSTM has controlled cell consisting of input,  forget, and output gate. These gates determine how  much previous  information  should  be  remembered  and  used  to predict the current state. In this paper, we added LSTM network after the CNN model to train sequential video frames. 
% 

%%%%%%%%%%%%%%%%%%%%%%%%%%%%%%%%%%%%%%%%%%%%%%%%%%%%%%%%%%%%%%%%%%%%%%%%%%%%%%%%
\section{Methodology}
The proposed framework is a multi-channel neural network that simultaneously and temporally assesses neonatal pain from face and body regions. Prior to pain assessment using our proposed framework, face and body regions detection should be performed. Neonatal face detection in the NICU is a challenging task due to several external factors such as illumination variations and partial occlusions (e.g., tapes or pacifier). As discussed in \cite{zamzmi2018review}, several facial landmark trackers faced difficulty when applied to detect and track the face of infants.  

\begin{figure*}[t]
		\begin{center}
		\includegraphics[width=\textwidth]{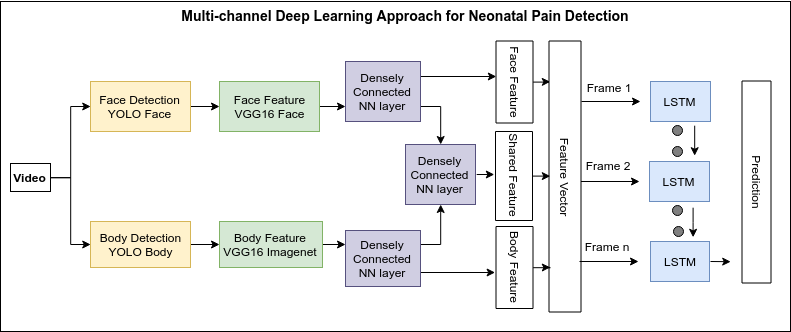}
		\caption{Overview of the proposed temporal and multi-channel network for assessing neonatal pain}
		\label{fig_framework}
		\end{center}
	\end{figure*}

In this paper, we used two pre-trained YOLO models to detect the face and body of neonates. The first YOLO detector was trained on WIDER Face \cite{yang2016wider} dataset to detect the faces of neonates. The second detector was trained on COCO \cite{lin2014microsoft} dataset to detect the body of neonates. After detecting the face and body regions from the input image, the proposed multi-channel neural network was used to extract features from face, body, and from both face and body together (shared representation). This architecture allows to extract shared features from both face and body in addition to the features extracted from them individually. After extracting the features from each frame, an LSTM model is used to capture the temporal information over the frames. The details of the framework are shown in Figure \ref{fig_framework}.

\subsection{Mutli-channel Shared Network}
Different pain scales such as NIPS \cite{hudson2002validation} and NPASS \cite{hummel2008clinical} are used to measure the neonatal pain in the NICU environment. Every scale provides a score for each pain channel or indicator such as the face, body, sound, and physiological signals. The final pain label should be decided by taking into account the scores or labels of all pain channels/indicators. Although pain assessment can be performed using a single channel, using multiple channels provides better pain assessment. This can be attributed to the fact that pain and other human emotions are expressed through multiple channels (e.g., see Figure \ref{fig_painFaceBody}). Therefore, we believe investigating the correlation between different pain channels and assessing neonatal pain using these channels together is important. 

\begin{figure}[t]
		\begin{center}
		\includegraphics[width=0.45\textwidth]{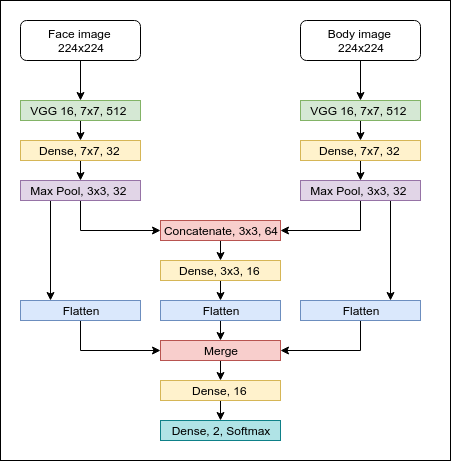}
		\caption{Overview of the multi-channel network for feature generation}
		\label{fig_multichannel}
		\end{center}
	\end{figure}

The proposed multi-channel CNN network can extract deep features from face and body individually as well as shared deep features from both face and body. We fine-tuned the model by freezing the layers of the pre-trained VGG16 model. Overview of the multi-channel shared CNN network is presented in Figure \ref{fig_multichannel}. For this CNN network, all the layers of VGG16 were frozen and only the upper layers of the proposed model are fine-tuned using Adam optimizer with a learning rate of 0.0001 and a batch size of 16 \cite{kingma2014adam}. 

As shown in Figure \ref{fig_multichannel}, the multi-channel shared CNN network used two pre-trained VGG16 based models. The first model was trained on VGGFace2 \cite{cao2018vggface2} dataset and used to extract features from the face of neonates. The second model was trained on ImageNet \cite{deng2009imagenet} dataset and used to extract features from the body of neonates. Each VGG16 model gets an RGB image of size ($224 \times 224$) as input and extracts deep features from the input image using a set of convolutional and max-pooling layers until the last convolutional layer of VGG-16 CNN architecture. Note that the green layers in Figure 3 represent the last convolutional layer of VGG-16 CNN architecture. After extracting deep features from face and body individually using VGG-16 models, the extracted features are sent to a dense (FC) layer of size ($7 \times 7$, 32) followed by a max-pooling layer of size ($3 \times 3$, 32). 

The down-sampled representations from face branch and body branch are then concatenated together into a single feature vector, which represents the shared representation. This shared representation is sent to a dense layer and flattened. To get the body feature vector, the down-sampled body representation is flattened. Similarly, the down-sampled face representation is flattened to get the face feature vector. After generating the face vector, body vector, and shared representation vector, these vectors are merged together followed by a dense or fully connected layer. Finally, the last layer performs binary classification using Softmax. The initial total number of training parameters was 29,474,818. Performing fine-tuning by freezing the VGG16 network reduces the total number of training parameters from 29,474,818 to 45,442.

\subsection{Integrating Temporal Information Using LSTM}
The multi-channel shared CNN network presented above is used to extract features from individual video's frames. To incorporate temporal changes into the model, we added LSTM layer to our proposed network to capture the changes in pain over time. We added two LSTM layers followed by a dense layer. We conducted two experiments: frame-level classification and video-level classification. 

As for the frame-level classification, a sequence of frames is sent to the LSTM layers for frame-level training and prediction after extracting the deep features (feature length 720) from the merge layer of the multi-channel neural network (Section III.A) from each frame. Each sequence has 16 consecutive frames. For example, if the first sequence is $S_0 = \{f_1, f_2, ..., f_n\}$, then second sequence is $S_1 = \{f_2, f_3, ..., f_{n+1}\}$, where $f_n$ represents the $n$-th time frame. The label of the last frame for each sequence represents the final label of that sequence. Note that a total of 16 frames per sequence was determined empirically. 

For the video-level classification, we used the entire sequence of frames to predict the label of the given video. The LSTM model was designed using a sigmoid function to predict pain confidence. Experimentally, we found that sigmoid function works better with LSTM than the cross-entropy error. To train the LSTM model, we used Adam optimizer \cite{kingma2014adam} with a learning rate of 0.0001 and a training batch size of 16. The details of the RNN network are shown in Table \ref{tab_rnn_architecture}.

\begin{table}
\centering
\caption{Details of the RNN Architecture} 
\begin{tabular}{|c|}
    \hline
    LSTM Unit 16 \\
    Activation = Tanh, Recurrent Activation = Hard Sigmoid\\
    \hline
    LSTM Unit 16 \\
    Activation = Tanh, Recurrent Activation = Hard Sigmoid\\
    \hline
    Dense  16, Relu\\
    \hline
    Dense  16, Relu\\
    \hline
    Dense  2, Sigmoid\\
    \hline
    Total parameters = 49,841\\
    \hline
\end{tabular}
\label{tab_rnn_architecture}
\end{table}

%%%%%%%%%%%%%%%%%%%%%%%%%%%%%%%%%%%%%%%%%%%%%%%%%%%%%%%%%%%%%%%%%%%%%%%%%%%%%%%%

\begin{table*}[t]
\centering
\caption{Frame Level Performance of Neonatal Pain Assessment}
\begin{tabular}{|l|c|c|c|c|}
\hline
       Approach     & Channel & Preprocess Dependency & Accuracy (\%) & AUC   \\ \hline
    ResNet50 & Face & Facial Landmarks & 85.83 &  0.84 \\ \hline
    VGG-16 & Face & Facial Landmarks & 89.47 & 0.89\\ \hline
   Proposed CNN + LSTM & Face + Body & Object Detection & 91.41 &  0.89 \\ \hline
\end{tabular}
\label{tab_framelevel}
\end{table*}

\begin{table*}[t]
\centering
\caption{Video Level Performance of Neonatal Pain Assessment}
\begin{tabular}{|l|c|c|c|c|c|}
\hline
            Approach    & Channel & Preprocess Dependency & Accuracy (\%) & AUC \\ \hline
    Geometric + SVM  & Face & Facial Landmarks & 88.32 & 0.82    \\ \hline
    LBP-TOP + SVM & Face & Facial Landmarks & 88.87 & 0.89    \\ \hline
    Motion + KNN & Body & Body Detection & 84.64 & 0.77 \\ \hline
    Proposed CNN + LSTM & Face + Body & Object Detection & 92.48 &  0.90\\ \hline
\end{tabular}
\label{tab_videolevel}
\end{table*}

\section{Experimental Results and Discussion}
\subsection{Dataset}
The dataset used in this paper was collected in the NICU at Tampa General Hospital, FL, USA. The dataset has a total of 31 neonates with age ranges from 32-40 gestational weeks (Mean age = 35.9 GW). Among the neonates, 12\% was Asian, 19\% was African American, 43\% was White, and 26\% was Caucasian. 

All the neonates were recorded using a GoPro camera while undergoing different procedural painful procedures such as heel lancing and immunization. We divided the data recording into eight-time periods. These periods are: the baseline period (T0), the procedure preparation period (T1), the painful procedure period (T2), and post-painful procedure periods (T3 to T7). The ground truth labels for each period was provided by two NICU using NIPS \cite{hudson2002validation} pain scale.

NIPS scale provides a pain score of 0 or 1 for face, body, and vital signs and a score of 0, 1, or 2 for crying sound. To get the final pain label, the scores from all modalities are added together followed by thresholding; i.e., a total score of 0-2 represents no-pain, a total score of 3-4 represents moderate, and a total score \textgreater 4 represents severe pain. The Kappa coefficient and Pearson correlation of the agreement between two caregivers were 0.85 and 0.89, respectively. 
%In total, the dataset has 184 videos recorded from 31 subjects. 
% (139 no-pain label, 37 pain label, and 7 moderate pain label).

Before proceeding further, it is worth to mention that COPE dataset \cite{brahnam2007introduction}, which consists of 204 static images of 26 neonates, is the only dataset we are aware of that is publicly available for research use. We have not used this dataset for building or evaluating the proposed framework since COPE images show only the neonates' face and the body region is wrapped with a blanket. Our proposed approach requires the presence of both face and body as inputs. 

\subsection{Preprocessing \& Evaluation Protocol}
We evaluated the performance of the proposed framework using the aforementioned dataset. We performed two experiments: frame-level classification and video-level classification. Prior to running the experiments, we converted all the videos in the dataset to 5 fps for evaluation. Because the length of the labeled videos in the utilized dataset is 10 seconds, we used 10 seconds video sequences. In this paper, we did not perform any level of augmentation. 

YOLO object detectors were then applied to each video's sequence to obtain the face and body regions in each frame. We then re-sized the images to $224 \times 224$ to accommodate with CNNs image’s size requirement of VGG-16 model (244 x 224 x 3, RGB images). In our experiments, we used leave-one-subject-out cross validation protocol for evaluation. 

\subsection{Results Analysis}
Table \ref{tab_framelevel} and Table \ref{tab_videolevel} present the experimental results of the proposed framework for frame-level and video-level, respectively. We compared the performance of our framework with the performance of similar existing works \cite{Zamzmi2019TBIOM, Zamzmi2019ITAC}, which used the same dataset presented in Section IV.A. 

Table \ref{tab_framelevel} presents the results of our frame-level classification and comparison with existing work \cite{Zamzmi2019TBIOM}. In \cite{Zamzmi2019TBIOM}, Zamzmi et al. used the popular ResNet50 and VGG16 architectures to classify static images or frames as pain or no-pain. Our proposed framework achieved 91.41\% average accuracy and 0.89 AUC. As shown in the table, the AUC of VGG-16 \cite{Zamzmi2019TBIOM} is the same as of the proposed framework. However, the proposed framework achieved better accuracy. 
% Although the performance of VGG-16 \cite{Zamzmi2019TBIOM} is similar to the performance of the proposed framework, the difference between their performance is not statistically significant. In addition, the proposed framework and VGG-16 has the same AUC value, which represents the performance of the pain class. 
Moreover, the proposed framework was trained using a smaller number of parameters 
%(VGG16 \cite{Zamzmi2019TBIOM}: 27,823,425 and this work: 45,442) 
and does not depend on facial landmark points. As reported in \cite{Zamzmi2019TBIOM}, the authors had to manually label facial landmark points in the failure frames. Usually, any facial tracker highly relies on face detection whereas YOLO based method works with different bounding boxes with confidence. Based on our experiments, we observed that YOLO detector performs better than facial tracker in several cases and has less time complexity.

Table \ref{tab_videolevel} shows the performance of the proposed framework in video-level classification as well as a comparison with existing works \cite{Zamzmi2019ITAC}. Instead of using deep learning based approach, the state-of-the-art handcrafted methods were used to perform video-level classification in \cite{Zamzmi2019ITAC}. As it can be seen from the table, the proposed framework achieved better average accuracy (92.48\%) and AUC (0.90) than three handcrafted methods used in \cite{Zamzmi2019ITAC}, namely geometric and LBP-TOP methods for face analysis and motion based method for body analysis. Moreover, the performance of existing works highly depends on the performance of the facial landmarks detected from the face.
The overall performance of the proposed framework is better than existing works which were evaluated on the same dataset. We believe this performance can be further improved with proper data augmentation. These results are encouraging and prove the feasibility of using the proposed framework for assessing neonatal pain. 

%%%%%%%%%%%%%%%%%%%%%%%%%%%%%%%%%%%%%%%%%%%%%%%%%%%%%%%%%%%%%%%%%%%%%%%%%%%%%%%%

\section{Conclusion and Future Work}
Accurate neonatal pain assessment is difficult, yet, very important task because inaccurate assessment can lead to over- or under- treatment. Due to partial or complete occlusion (e.g., medical tapes or oxygen mask), illumination variations, immature facial muscles, detecting and tracking facial landmark points can become very challenging. In this paper, a facial-landmarks-free deep learning based framework is proposed to assess neonatal pain from videos. The proposed framework, which is called temporal and multi-channel network, integrates multiple channels (face and body) and temporal information using LSTM to investigate the correlation between different channels over time. Preliminary experimental results demonstrate the feasibility of using the proposed framework for assessing neonatal pain. 

In the future, we plan to follow four main directions. First, we plan to perform spatial (frame-level) and temporal (sequence-level) augmentation to enlarge the training set. Second, we want to explore methods for increasing the number of the minor class to get a relatively balanced dataset. Third, we plan to integrate other pain modalities such as crying sound and vital signs into the current framework. Finally, we would like to evaluate the proposed framework on a dataset collected from neonates while they were experiencing postoperative pain. At present, we have a small postoperative dataset collected from 9 subjects, which can be used for evaluation, but not enough for training. We are currently involved in an ongoing effort to collect a larger pain dataset from neonates during postoperative pain.


%%%%%%%%%%%%%%%%%%%%%%%%%%%%%%%%%%%%%%%%%%%%%%%%%%%%%%%%%%%%%%%%%%%%%%%%%%%%%%%%

\section*{ACKNOWLEDGMENT}
We are grateful for the parents who allowed their babies to take part in this study and the entire neonatal staff at TGH.

%%%%%%%%%%%%%%%%%%%%%%%%%%%%%%%%%%%%%%%%%%%%%%%%%%%%%%%%%%%%%%%%%%%%%%%%%%%%%%%%

% \begin{thebibliography}{99}
% \end{thebibliography}

\bibliographystyle{IEEEtran}
\bibliography{bibliography}
%%%%%%%%%%%%%%%%%%%%%%%%%%%%%%%%%%%%%%%%%%%%%%%%%%%%%%%%%%%%%%%%%%%%%%%%%%%%%%%%

\end{document}